\documentclass[twoside,11pt]{article}

\usepackage{melba}

\usepackage{amsmath,amsfonts}

\usepackage{microtype}
\usepackage{graphicx}
\usepackage{subfigure}
\usepackage{booktabs} 
\usepackage{cleveref}
\usepackage{float}
\usepackage{enumitem}
\usepackage{listings}

\newcommand{\IN}{\emph{In} }
\newcommand{\OUT}{\emph{Out} }
\usepackage{multirow, makecell}
\usepackage{multicol}
\usepackage{pifont}
\melbaheading{1}{2020}{1-48}{10/2019}{1/2020}{Cao, Huang, Hui, Cohen}

\ShortHeadings{A Benchmark of Medical Out of Distribution Detection}{Cao, Huang, Hui, Cohen}
\firstpageno{1}

\title{A Benchmark of Medical Out of Distribution Detection}

\author{\name Tianshi Cao \email jcao@cs.toronto.edu \\  
	\addr Vector Institute, University of Toronto
	\AND
	\name Chin-Wei Huang \email cw.huang427@gmail.com \\
	\addr Mila, University of Montreal
	\AND
	\name David Yu-Tung Hui \email dythui1@gmail.com \\
	\addr Mila, University of Montreal
	\AND
	\name Joseph Paul Cohen \email joseph@josephpcohen.com \\
	\addr Mila, University of Montreal
}

\editors{Tal Arbel, more names}

\begin{document}

\maketitle

\begin{abstract}
\noindent \textbf{Motivation:}
Deep learning models deployed for use on medical tasks can be equipped with Out-of-Distribution Detection (OoDD) methods in order to avoid erroneous predictions. However it is unclear which OoDD method should be used in practice. 

\noindent \textbf{Specific Problem:}
Systems trained for one particular domain of images cannot be expected to perform accurately on images of a different domain.  These images should be flagged by an OoDD method prior to diagnosis. 

\noindent \textbf{Our approach:}
This paper defines 3 categories of OoD examples and benchmarks popular OoDD methods in three domains of medical imaging: chest X-ray, fundus imaging, and histology slides. 

\noindent \textbf{Results:}
Our experiments show that despite methods yielding good results on some categories of out-of-distribution samples, they fail to recognize images close to the training distribution.

\noindent \textbf{Conclusion:}
We find a simple binary classifier on the feature representation has the best accuracy and AUPRC on average. Users of diagnostic tools which employ these OoDD methods should still remain vigilant that images very close to the training distribution yet not in it could yield unexpected results.
\end{abstract}

\section{Introduction}
A safe system for medical diagnosis should withhold diagnosis on cases outside its validated expertise. For machine learning (ML) systems, the expertise is defined by the validation score on the distribution of data used during training, as the performance of the system can be validated on samples drawn from the same distribution (as per PAC learning \citep{Valiant1984}). 
This restriction can be translated into the task of \emph{Out-of-Distribution Detection} (OoDD), the goal of which is to distinguish between samples in and out of the training distribution of the diagnosis system (abbreviated to \IN and \OUT data). 

In contrast to natural image analysis, medical image analysis must often deal with orientation invariance (e.g. in cell images), high variance in feature scale (in X-ray images), and locale specific features (e.g. CT) \citep{Razzak_2017}.  
A systematic evaluation of OoDD methods for applications specific to medical image domains remains absent, leaving practitioners blind as to which OoDD methods perform well and under which circumstances.
This paper fills this gap by benchmarking many current OoDD methods under four medical image types (frontal and lateral chest X-ray, fundus imaging, and histology). Our empirical studies show that current OoDD methods perform poorly when detecting correctly acquired images that are not represented in the training data. We also find that some simple methods such as a binary classifier on features trained on \IN data performed on par with more complex methods (see Figure~\ref{fig:all_acc}). We hope that this work can inspire more discussion and future work on the unique challenges of OoDD in medical image domains.

\begin{figure}[t]
    \centering
    \vspace{-15pt}
    \includegraphics[width=0.9\columnwidth]{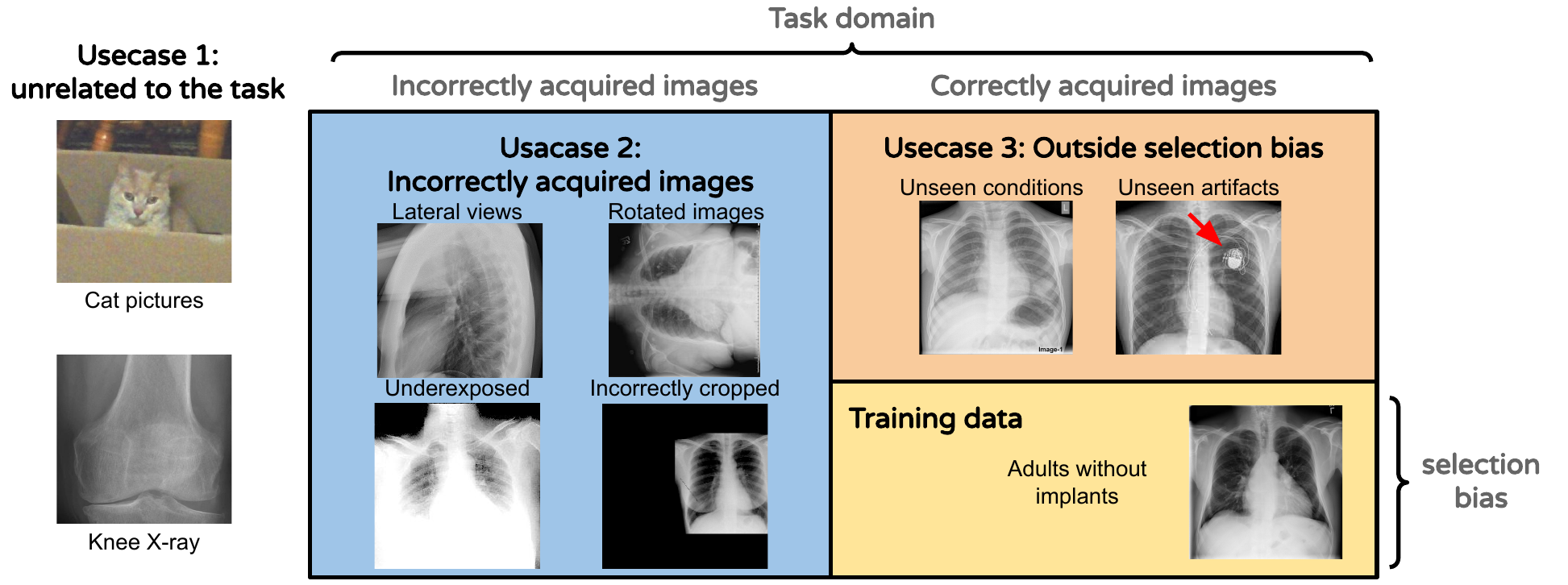}
    \vspace{-15pt}
    \caption{The three use-cases shown in relation to each other. The training data is sampled iid from the \IN data distribution. 1. Inputs that are unrelated to the task. 2. Inputs which are incorrectly prepared 3. Inputs that are unseen due to a selection bias in the training distribution.}
    \label{fig:venn}
    \vspace{-15pt}
\end{figure}

\section{Defining OoD in Medical Data}\label{sec:ood}
Given an \IN distribution dataset, how should we define what constitutes \OUT data? To address this, we identify three distinct out-of-distribution categories:
\begin{itemize}
    \item \textbf{use-case 1} Reject inputs that are unrelated to the evaluation. This includes obviously-wrong images from a different domain (e.g. MRI images processed using a model trained on X-ray images) and less obviously-wrong images (e.g. wrist X-ray image processed using a model trained with chest X-rays).
    \item \textbf{use-case 2} Reject inputs which are incorrectly prepared For example, in the case of chest X-ray images: blurry images, poor contrast, incorrect view of the anatomy (lateral views processed using a model trained with frontal views), images with the incorrect file format or pre-processing applied), or changes in data acquisition protocol.
    \item \textbf{use-case 3} Reject inputs that are unseen due to a selection bias in training data (e.g. image with an unseen disease), which may yield unexpected results.
\end{itemize}

We justify these use-cases by enumerating different types of mistakes or biases that can occur at different stages of the data acquisition. This is visually represented in Figure \ref{fig:venn}. We construct our experiments to evaluate OoDD methods' performance on each category.

\paragraph{Example 1} As running example, we will use our first evaluation where the \IN data consists of frontal chest X-rays. The \IN data contains 10 pulmonary conditions in the NIH Chest X-ray dataset \citep{WangNIH2017}. In use-case 1 we include natural images, images of symbols and text, and skeletal X-ray images. use-case 2 contains lateral view, dorsal view, and pediatric chest x-rays. Finally, use-case 3 include frontal chest X-rays of four pulmonary conditions that were not present in \IN data.

\section{Task Formulation} \label{sec:task}
\begin{figure}[t]
    \centering
    \includegraphics[width=\textwidth]{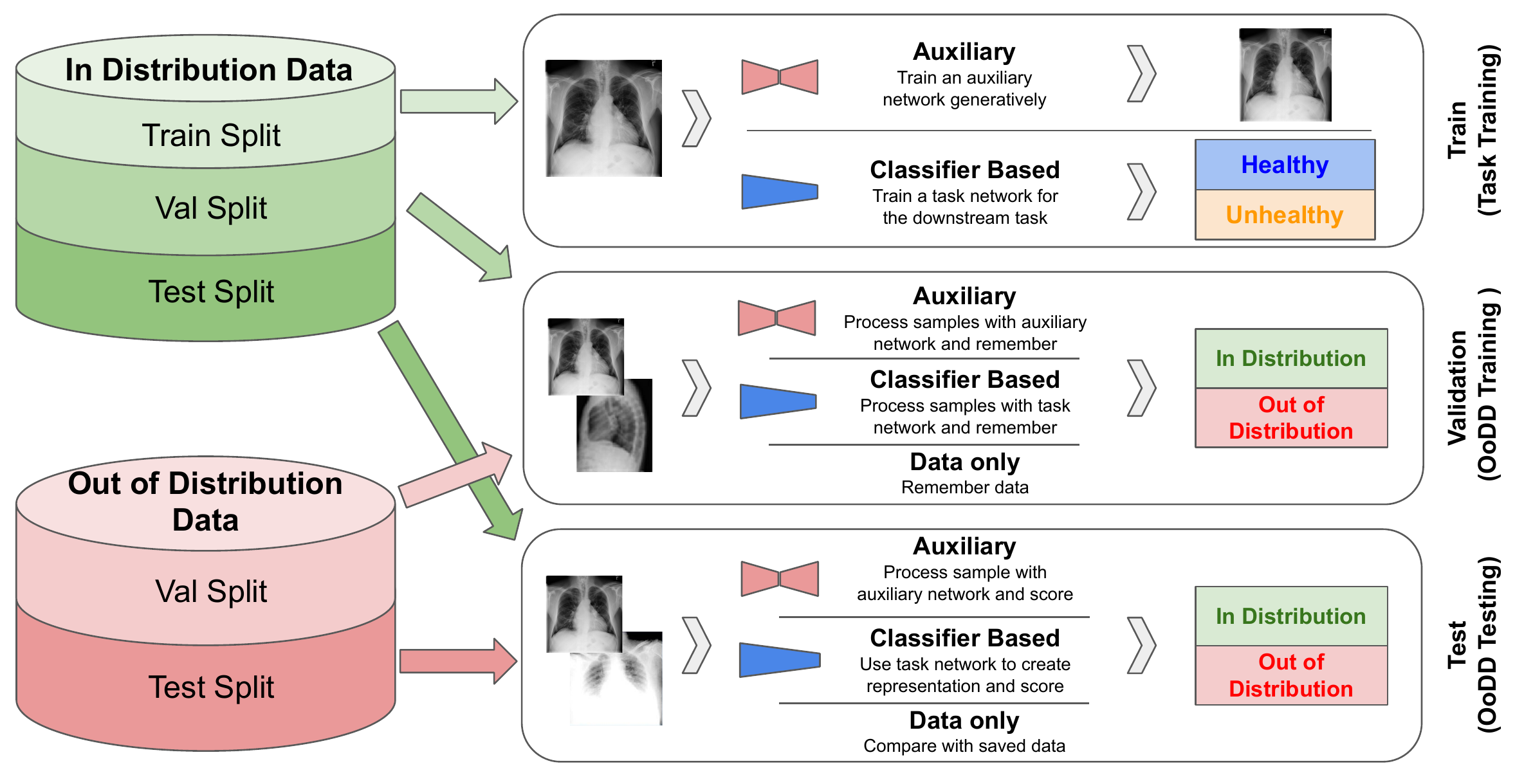}
    \vspace{-20pt}
    \caption{Interplay of In and Out Data with three steps of OoDD evaluation. The data splits are shown on the left for the \IN and \OUT data. On the right, three parts of the evaluation are shown. At the top the classifier or auxiliary network is trained. The OoDD method is trained using validation data in the middle and then evaluated on test data in the bottom.}
    \label{fig:method}
    \vspace{-20pt}
\end{figure}

In this paper, we will assume that the downstream task is to perform classification using a deep neural network, which we call the task network. Let us denote a sample of \IN data used to train the task network as $D_{tr}$. Auxiliary models, as required by some OoDD methods, are also be trained on $D_{tr}$. Then, an OoDD method $M$ is trained on a ``validation set" $D_{val} = D_{val}^{in} \cup D_{val}^{out}$, a union of \IN and \OUT samples (labeled as ``in" or ``out"). $M$ may also use the features learned by the task network, thereby also making use of $D_{tr}$. Finally, $M$ is evaluated on the test set $D_{test} = D_{test}^{in} \cup D_{test}^{out}$, also composed of \IN and \OUT samples. Each tuple $(M, D_{tr}, D_{val}^{in},D_{val}^{out} , D_{test}^{in}, D_{test}^{out})$ constitutes an experiment. This three step process is illustrated in figure \ref{fig:method}.

\subsection{Methods of OoDD (\texorpdfstring{$M$}{M})}\label{sec:methods}
\label{sec:MOOD}
We consider three classes of OoDD methods. Data-only methods do not rely on any pre-trained models and are learned directly on $D_{val}$. Classifier-only methods assume access to a downstream classifier trained for classification on \IN data ($D_{tr}$). Methods with auxiliary models requires pre-training of a neural network that on \IN data using other objectives such as image reconstruction. 

\paragraph{Data-only methods}
The most simple and easy to implement data-only baseline is k-Nearest-Neighbors (KNN) which only needs to observe the training data. This is performed on images as a baseline for our evaluations. For speed only 1000 samples are used from $D_{tr}$ to calculate neighbor distance. A threshold is determined using samples from $D_{val}$.

\paragraph{Classifier-only methods}
Classifier-only methods make use of the downstream classifier for performing OoDD. Compared to data-only methods they require less storage, however their applicability is constrained to cases with classification as downstream tasks.
\emph{Probability Threshold} \citep{Hendrycks2017} uses a threshold on the prediction confidence of the classifier to perform OoDD. 
\emph{Score SVM} trains an SVM on the logits of the classifier as features, generalizing probability threshold.
\emph{Binary Classifier} trains on the features of the penultimate layer of the classifier. 
\emph{Feature KNN} uses the same features as the binary classifier, but constructs a KNN classifier in place of logistic regression.
\emph{ODIN} \citep{Liang2017} is a probability threshold method that preprocesses the input by taking a gradient step of the input image to increase the difference between the \IN and \OUT data.
\emph{Mahalanobis} \citep{Lee2018a} models the features of a classifier of \IN data as a mixture of Gaussians, preprocesses the data as ODin, and thresholds the likelihood of the feature. 

\paragraph{Methods with Auxiliary Models}
OoDD methods in this section require an auxiliary model trained on \IN data. This results in extra setup time and resources when the downstream classifier is readily available. However, this could also be advantageous when the downstream task is not classification (such as regression) where methods may be difficult to adapt.
\emph{Autoencoder Reconstruction} thresholds the reconstruction loss of the autoencoder to achieve OOD detection. 
Intuitively, the autoencoder is only optimized for reconstructing \IN data, and hence reconstruction quality of \OUT data is expected to be poor due to the bottleneck in the autoencoder. 
In this work we consider three variants of autoencoders: standard autoencoder (AE) trained with reconstruction loss only, variational autoencoder trained with a variational lower bound (VAE) \citep{Kingma2013}, and decoder+encoder trained with an adversarial loss (ALI \citep{Dumoulin2016}, BiGAN \citep{Donahue2017a}).
Furthermore, we include two different reconstruction loss functions in the benchmark: mean-squared error (MSE) and binary cross entropy (BCE). 
Finally, \emph{AE KNN} constructs a KNN classifier on the features output by the encoder.

\paragraph{Example 1 (cont.)} We will use \emph{Autoencoder Reconstruction with VAE trained using MSE Loss}(Reconst. VAEMASE) as the OoDD method of our running example. In the first stage, we train the auxiliary VAE on $D_{tr}$ by maximizing the evidence lower bound (ELBO) under MSE criteria as evidence. Then, in the second stage, we compute the reconstruction loss on samples of $D_{val}$ and calibrate a threshold value on reconstruction loss for separating \IN and \OUT samples. Finally, we evaluate on $D_{test}$ by predicting its label (``in" or ``out" ) according to the reconstruction loss and comparing to the ground truth. 

\subsection{\IN Datasets (\texorpdfstring{$D_{tr}, D_{val}^{in}, D_{test}^{in}$)}{D\_tr, D\_val\^in, D\_test\^in}}\label{sec:in_data}
\begin{table*}[bth]
    \centering
    \resizebox{\textwidth}{!}{\begin{tabular}{cccccr}
        \toprule
         Domain & Eval & \IN data & use-case 1 \OUT data & use-case 2 \OUT data & use-case 3 \OUT data  \\
         \midrule
         \multirowcell{4}{\mbox{Chest X-ray}} & \multirowcell{2}{\mbox{1}} & \multirowcell{2}{NIH \\ (\IN split)} & UC-1 Common, MURA & \multirowcell{2}{PC-Lateral, PC-AP, \\PC-PED, PC-AP-Horizontal} & 
         \multirowcell{2}{NIH-Cardiomegaly, NIH-Nodule,\\ NIH-Mass, NIH-Pneumothorax }\\
         & & & & \\
         \addlinespace
         & \multirowcell{2}{\mbox{2}} & 
         \multirowcell{2}{PC-Lateral \\ (\IN split)} &  UC-1 Common, MURA & \multirowcell{2}{PC-AP, PC-PED, \\ PC-AP-Horizontal, PC-PA} &  \multirowcell{2}{PC-Cardiomegaly, PC-Nodule,\\ PC-Mass, PC-Pneumothorax } \\
         & & & & \\
         \midrule
         \addlinespace
         Fundus Imaging & 3 & DRD & UC-1 Common & DRIMDB & RIGA \\
         \midrule
         \addlinespace
         Histology & 4 & PCAM & UC-1 Common, Malaria & ANHIR, IDC & None \\
        \bottomrule
    \end{tabular}}
    \caption{\small Datasets used in evaluations. UC-1 Common includes datasets such as MNIST, CIFAR-10, and random noise. PC=PadChest, NIH=NIH chest X-ray8, DRIMDB=Diabetic Retinopathy Images Database, RIGA=Retinal fundus images for glaucoma analysis. See Appendix \ref{sec:dataselection} for more details.}
    \label{tab:dataset_roster}
\end{table*}
For $D_{tr}$, we select from four medical datasets ranging over three modalities of medical imaging. Each dataset has been randomly split three ways for use in $D_{tr}$, $D_{val}^{in}$, and $D_{test}^{in}$. Each dataset also contains a classification task. As most ML applications only deal with one image type (i.e. an medical application wouldn't simultaneous diagnose chest conditions and diabetic retinopathy), we consider each \IN distribution dataset as distinct evaluations and do not consider their combinations. The \IN datasets of each evaluation are:
\begin{enumerate}
    \item Frontal view chest X-ray images. The task is to predict 10 of the 14 radiologcal findings defined by the \textbf{NIH} Chest-X-Ray dataset \citep{WangNIH2017}. The remaining conditions are held-out for use-case 3.
    \item Lateral view chest X-ray images (PC-Lateral). The task is the same as evaluation 1, but the data is from lateral view images in the PADChest (\textbf{PC}) dataset \citep{Bustos2019}. Remaining conditions are also held-out for use-case 3.
    \item Fundus/retinal (back of the eye) images. The task is to detect diabetic retinopathy in the retina defined by the \textbf{DRD} (Diabetic Retinopathy Detection) dataset. \citep{kaggle-diabetic-retinopathy}
    \item H\&E stained histology slides of lymph nodes. The task is to predict if image patches contain cancerous tissue defined by the \textbf{PCAM} dataset \citep{Veeling2018}. 
\end{enumerate}

\subsection{\OUT Datasets (\texorpdfstring{$D_{val}^{out}$}{D\_val\^out} and \texorpdfstring{$D_{test}^{out}$}{D\_test\^out})}\label{sec:out_data}
We select \OUT datasets according to use-cases described in \cref{sec:ood}. As users may be independently interested in a particular use-case, we evaluate the OoDD methods per use-case. Clearly, characteristics of each use-case are defined relative to the \IN distribution, hence we may need to select different \OUT datasets for each \IN dataset.

For $D_{val}^{out}$ and $D_{test}^{out}$ under \textbf{use-case 1}, we take a combination of natural image and symbols datasets which we call \emph{UC-1 Common}. This is used for every \IN data. For \textbf{use-case 2}, we use datasets of the same modality of the \IN distribution, but incorrectly captured. For example, different views (e.g. lateral vs frontal) of the chest area are used as $D_{val}^{out}$ and $D_{test}^{out}$ for evaluations 1 and 2. Finally, for \textbf{use-case 3}, we use images of different conditions/diseases as \OUT data. For evaluations 1 and 2, the four held-out conditions are used as use-case 3 \OUT data. We did not include a use-case 3 \OUT dataset for histology slides due to lack of available data. Table \ref{tab:dataset_roster} summarizes our roster of \IN and \OUT datasets. Each \OUT dataset is split 50/50 for $D_{val}^{out}$ and $D_{test}^{out}$. Subsampling is used to balance the number of \IN and \OUT samples in $D_{val}$ and $D_{test}$.

It remains to be determined how to split \OUT data between $D_{val}$ and $D_{test}$. A common but overly optimistic assumption is that \OUT data are similar to each other, hence the OoDD method is trained and evaluated on different splits of the same OoD dataset. In our running example, this entails calibrating the threshold for reconstruction loss on NIH Chest data vs MNIST training-split, and then evaluate on NIH chest data vs MNIST testing split. On the other extreme, the assumption is that we have no access to out-of-distribution data, turning the task into that of one-class classification where no \OUT data is used except for testing. In a realistic setting, the developer would train the OoDD method on a number of various datasets to cover different modes of OoD data, but the data seen at deploy time possesses variability not accounted for by those selected by the developer. Hence, for each use-case, we select a subsample of datasets for training the OoDD method, and use the remaining datasets for evaluation.

\paragraph{Example 1. (cont.)} For use-case 1 of the running example, we split the \OUT data in to 14 partitions (9 datasets in UC-1 Common, and 5 areas of the body in the MURA skeletal X-ray dataset). We sample without replacement 3 partitions for $D_{val}^{out}$, and use the rest in $D_{test}^{out}$. In use-case 2, we have lateral-view, pediatric (PED), dorsal-view (AP), and horizontal dorsal-view (AP-Horizontal) as four \OUT splits. We randomly select one as $D_{val}^{out}$ and use the remaining for $D_{test}^{out}$. We do the same for use-case 3, which also has four \OUT splits.

\begin{figure*}[bht]
    \centering
    \includegraphics[width=\textwidth]{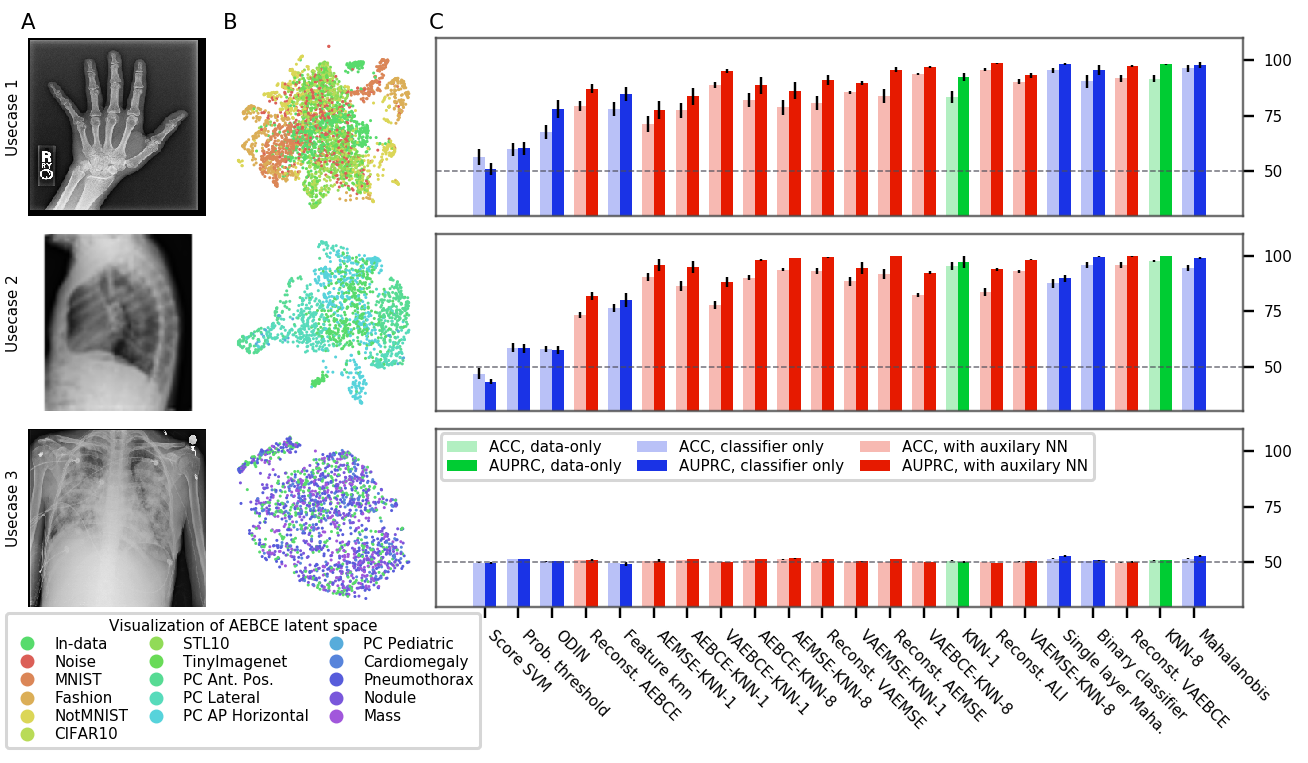}
    \vspace{-20pt}
    \caption{\small Visualizations and OoDD results on AP view chest-xray (Evaluation 1). Each row of figures correspond to a use-case. Column A shows examples of \OUT data for each use-case (hand x-ray, lateral view chest X-ray, and xray of pneumothorax from top to bottom). Column B shows UMAP visualizations of AE latent space - colors of points represent their respective datasets. Column C plots the accuracy and AUPRC of OoDD methods in each use-case, averaged across all randomized trials. Bars are sorted by accuracy averaged across use-cases, and coloured according to method's grouping: green for baseline image space methods, blue for methods based upon the task specific classifier, and red for methods that use an auxilary neural network. Error bars represent $95\%$ confidence interval.}
    \label{fig:nih_mainfig}
\end{figure*}

\section{Experiments and Results} \label{sec:results}
In this benchmark, we report the performance of each OoDD method on every evaluation and use-case. We measure the accuracy and Area Under Precision-Recall Curve (AUPRC) on $D_{test}$, totaling at 11 pairs of performance numbers per method. Since $D_{test}$ is class-balanced, accuracy provides an unbiased representation of type I and type II errors. AUPRC characterizes the separability of \IN and \OUT samples in predicted value (the value that we threshold to obtain classification). Details of experimental setup are in Appendix \ref{sec:procedure}.

\begin{figure*}[t]
    \centering
    \includegraphics[width=0.95\textwidth]{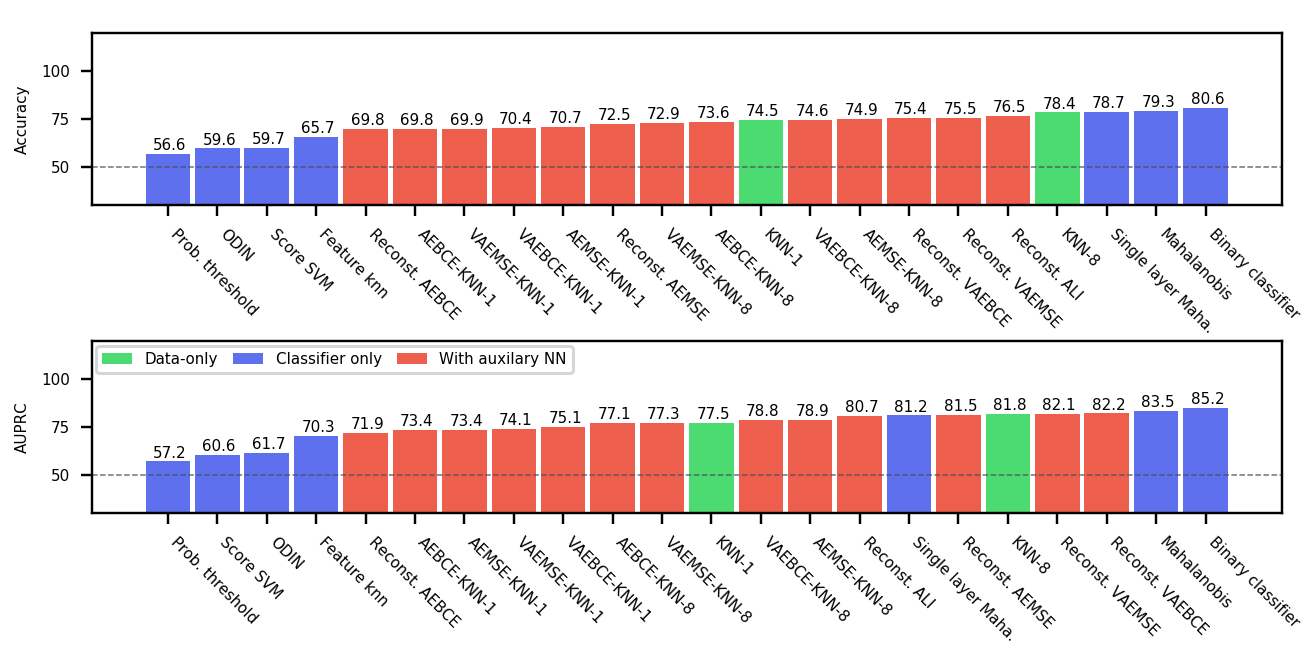}
    \vspace{-5pt}
    \caption{Accuracy and AUPRC of OoDD methods aggregated over all evaluations. Sorted by accuracy from left to right.}
    \label{fig:all_acc}
\end{figure*}

\begin{figure*}[t]
    \centering
    \includegraphics{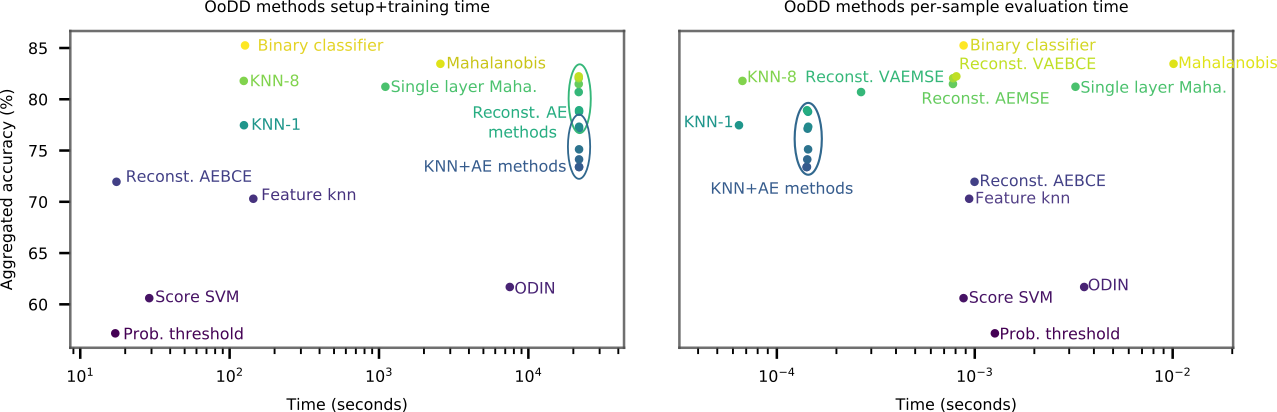}
    \vspace{-5pt}
    \caption{Overall accuracy of methods plotted over total setup time (left) and per-sample run time (right)}
    \label{fig:time_v_perf}
\end{figure*}

\begin{figure*}
    \centering
    \includegraphics[width=\textwidth]{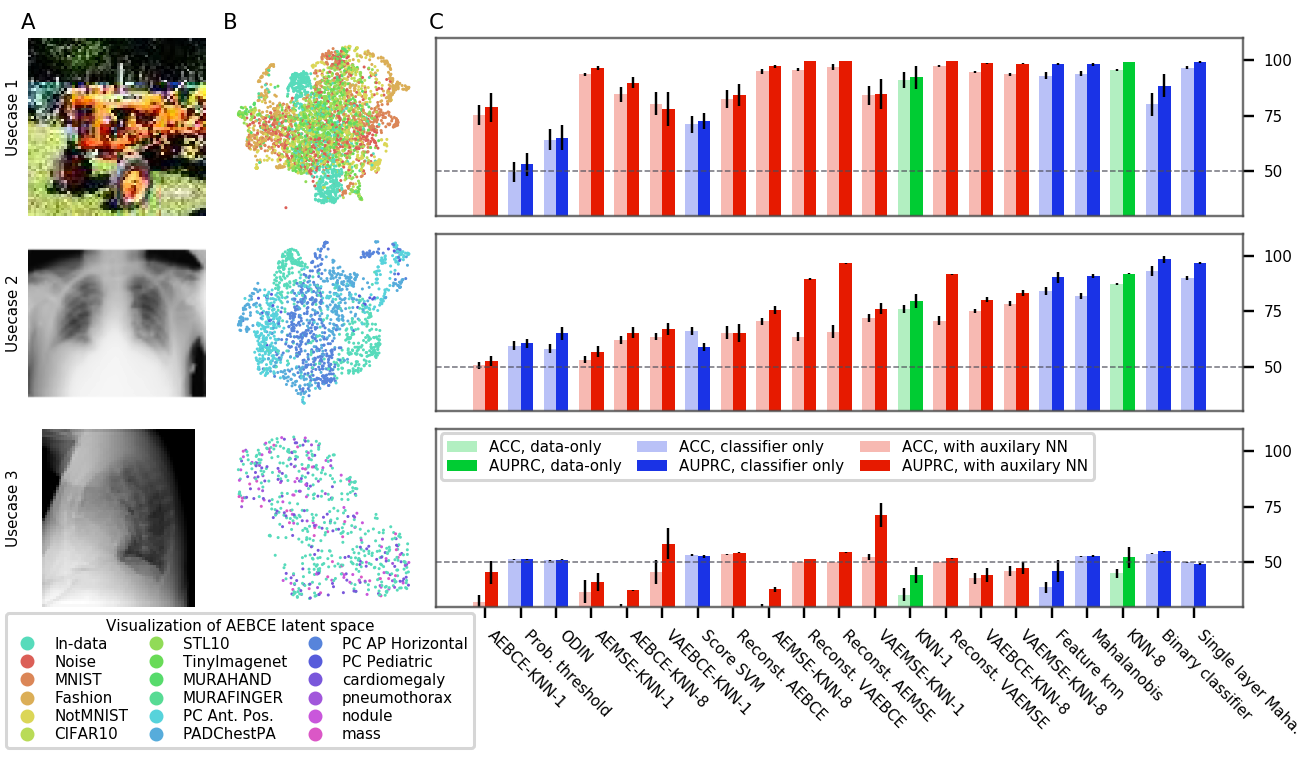}
    \vspace{-20pt}
    \caption{Lateral X-ray imaging (see Figure 
    \ref{fig:nih_mainfig} for description)}
    \label{fig:pad_mainfig}
\end{figure*}

\begin{figure*}
    \centering
    \includegraphics[width=\textwidth]{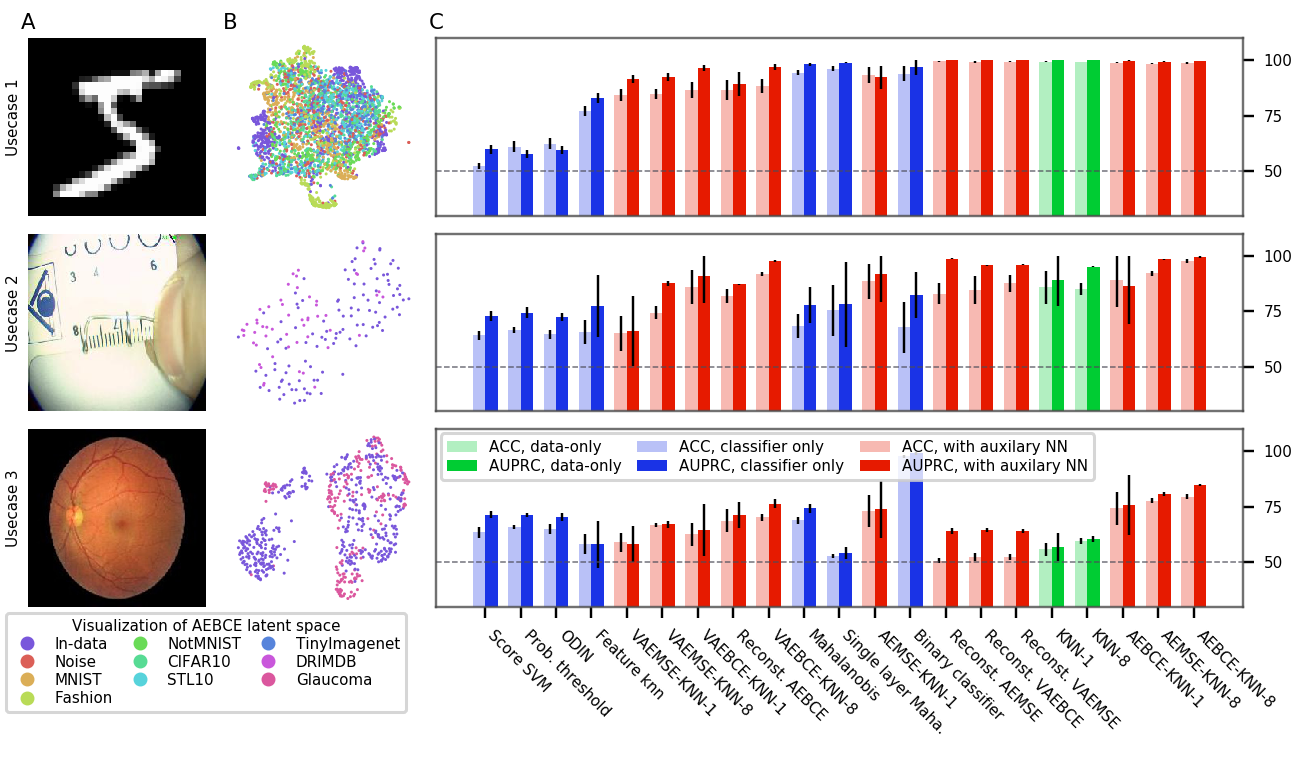}
    \vspace{-30pt}
    \caption{Fundus Imaging (see Figure 
    \ref{fig:nih_mainfig} for description)}
    \label{fig:drd_mainfig}
\end{figure*}

\begin{figure*}
    \centering
    \includegraphics[width=\textwidth]{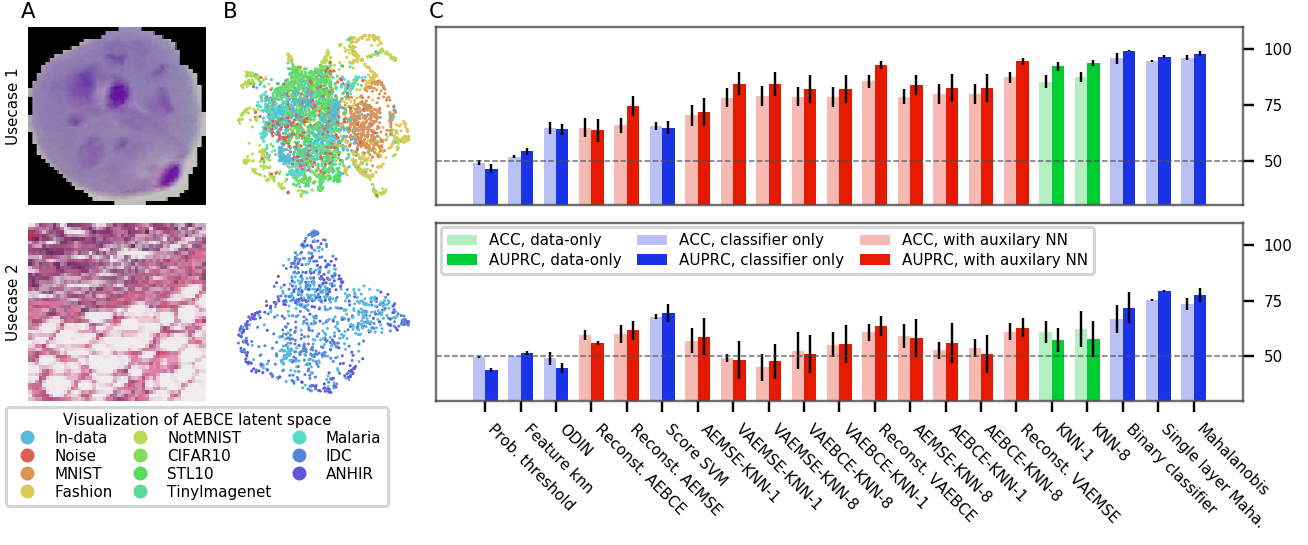}
    \vspace{-30pt}
    \caption{Histology Imaging (see Figure 
    \ref{fig:nih_mainfig} for description)}
    \label{fig:pcam_mainfig}
\end{figure*}

Figures \ref{fig:nih_mainfig}, \ref{fig:pad_mainfig}, \ref{fig:drd_mainfig}, and \ref{fig:pcam_mainfig} show the performance of OoDD methods on the four evaluations. 
Generally, we observe that our choice of datasets create a range of simple to hard test cases for OoDD methods. 
While many methods can solve use-case 1 and use-case 2 adequately in evaluations 1-3, use-case 3 proves difficult for all methods tested. 
This is reflected in the UMAP visualization of the AE latent spaces (column B of figures \ref{fig:nih_mainfig} to \ref{fig:drd_mainfig}), in which we observe that the \IN data points are easily separable from \OUT data in use-cases 1 and 2, but well-mixed with \OUT data in use-case 3. 
It is surprising that no method achieved significantly better accuracy than random in use-case 3 of evaluations 1 and 2 across all repeated trials. 
This illustrates the extreme difficulty of detecting unseen/nouveau diseases, which corroborates the findings of \citet{ren2019likelihood}. 

\textbf{Overall Performance} Across evaluations, the better performing classifier-only methods are competitive with the methods that use auxiliary models. When performance is aggregated across all evaluations (Figure \ref{fig:all_acc}), the best classifier-only methods (Mahalanobis and binary classifier) outperform auxiliary models in accuracy. The performance of binary classifier is strong despite the method's simplicity. We suspect that this strong performance is due to the fact that we randomly sample 3 \OUT datasets when constructing $D_{val}$ as opposed to selecting a single \OUT dataset. This added variety in $D_{val}$ \OUT data improves generalization by enforcing more stable decision boundaries. We performed additional experiments with fewer \OUT datasets on a subset of methods and tasks. Results in appendix figure \ref{fig:top4_perf} shows that the gap between the top-4 methods quickly closing with more \OUT datasets in $D_{val}$.

\paragraph{Computational Cost}
We consider computational cost of each method in terms of setup time and run time. The setup time is measured as the wall-clock computation time taken for hyperparameter search and training. For methods with auxiliary models, the training time of auxiliary neural networks are also included in the setup-time. Run time is measured as the per-sample computation time (averaged over fixed batch size) at test time. Figure \ref{fig:time_v_perf} plots the accuracy of models over their respective setup and run time. All methods can make predictions reasonably fast, allowing for potential online usage. Mahalanobis and its single layer variant take significantly more time to setup and run than other classifier methods. KNN-8 exhibits the best time vs performance trade-off with its low setup time and good performance. However, as it requires the storage of training images for predictions, it may be unsuitable for use on memory constrained platforms (e.g. mobile) or when training data privacy is of concern.


\section{Discussion}
The necessity of OoDD is supported by two considerations. First of which is usability. As we transition Machine Learning tools from research labs to the hands of the end user, usability of these tools becomes pivotal to their success. One common characteristic of good usability is to fail gracefully when handling user errors. In ML assisted diagnostic tools, this means equipping the tool with the capacity to reject predictions on erroneous input data, thereby preventing the “garbage-in, garbage-out” scenario. For ML tools facing users outside the computer science community, this clarity is particularly important. The second reason why OoDD is necessary is the requirement for safety. In applications like ML assisted diagnosis, the performance of the system is directly tied to the safety of the patients. A well documented failure mode for Machine learned predictors is when the predictor attempts to extrapolate on inputs outside the distribution of its training data. OoDD provides a safety mechanism that prevents failures of the predictor from harming the user through inaccurate predictions.
\section{Conclusion}
Overall, the top three classifier-only methods obtain better accuracy than all methods with auxiliary models except for fundus imaging.
Binary classifier has the best accuracy and AUPRC on average, and is simple to implement. Hence, we recommend binary classifier as the default method for OoDD in the domain of medical images. The methods we find to work best are almost opposite that of \citet{Shafaei2018} despite using the same code for overlapping methods. The main difference between these studies is that they evaluate on natural images instead of medical images.  We performed an extensive hyperparameter search on all methods and conclude that this discrepancy is due to the specific data and tasks we have defined.
While use-case 1 and 2 are easily solved with non-complicated models, the failure of most models in almost all tasks to significantly solve use-case 3 is consistent with the finding of \citet{Ahmed2019}.  Users of diagnostic tools which employ these OoDD methods should still remain vigilant that images very close to the training distribution yet not in it (and a false negative for use-case 3) could yield unexpected results. In the absence of OoDD methods which have good performance on use-case 3, another approach is to develop methods which will systematically generalize to these examples.

\newpage
\acks{We thank Tobias Würfl, Faruk Ahmed, and Ronald Summers for their useful comments. This work utilized the supercomputing facilities managed by Compute Canada and Calcul Quebec. We thank AcademicTorrents.com for making data available for our research.}

\bibliography{neuralnetworks,other}
\newpage
\appendix
\section{Description of Datasets}\label{sec:dataselection}
The following datasets are used in UC-1 Common:
\begin{itemize}[nosep]
    \item \textbf{MNIST}\footnote{http://yann.lecun.com/exdb/mnist/} 28x28 black and white hand written digits data. Original test split is used in UC-1 Common. 
    \item \textbf{notMNIST} \footnote{http://yaroslavvb.blogspot.com/2011/09/notmnist-dataset.html} Letters A-J in various fonts. Black and white with resolution of 28x28. Test split is used.
    \item \textbf{CIFAR10 and CIFAR100}\footnote{https://www.cs.toronto.edu/~kriz/cifar.html} 32x32 natural images. Original test split used in UC-1 Common.
    \item \textbf{TinyImagenet}\footnote{https://tiny-imagenet.herokuapp.com/} 96x96 downsampled subset of ILSVRC2012. Validation split used in UC-1 Common.
    \item \textbf{FashionMNIST}\footnote{https://www.kaggle.com/zalando-research/fashionmnist} Grayscale 28x28 images of clothes and shoes. Validation split is used in UC-1 Common.
    \item \textbf{STL-10} \footnote{https://ai.stanford.edu/~acoates/stl10/} Natural image dataset of size 96x96. 8000 testing images are used in UC-1 Common.
    \item \textbf{Noise} White noise generated at any desired resolution.
\end{itemize}
    
The following medical datasets are used:
\begin{itemize}[nosep]
    \item \textbf{ANHIR} \footnote{https://anhir.grand-challenge.org/} Automatic Non-rigid Histological Image Registration Challenge.  Microscopy images of histopathology tissue samples stained with different dyes. Images of intestine and kidney tissue were used in evaluation 4, use-case 2.
    \item \textbf{DRD} \footnote{https://www.kaggle.com/c/diabetic-retinopathy-detection/data} High-resolution retina images with presence of diabetic retinopathy in each image labeled on a scale of 0 to 4. We convert this into a classification task where 0 corresponds to healthy and 1-4 corresponds to unhealthy. 
    \item \textbf{DRIMDB} Fundus images of various qualities labeled as good/bad/outlier. We use the images labeled as bad/outlier in evaluation 3, use-case 2.
    \item \textbf{Malaria} \footnote{https://lhncbc.nlm.nih.gov/publication/pub9932} Image of cells in blood smear microscopy collected from healthy persons and patients with malaria. Used in evaluation 4 use-case 1.
    \item \textbf{MURA} \footnote{https://stanfordmlgroup.github.io/competitions/mura/} MUsculoskeletal RAdiographs is a large dataset of skeletal X-rays. We use its validation split in evaluation 1 and 2's use-case 1. Images are grayscale and the square cropped.
    \item \textbf{NIH Chest} \footnote{https://www.kaggle.com/nih-chest-xrays/data} This NIH Chest X-ray Dataset is comprised of 112,120 X-ray images with 14 condition labels. The x-rays images are in posterior-anterior view (X-tray traverses back to front).
    \item \textbf{PAD Chest} \footnote{https://bimcv.cipf.es/bimcv-projects/padchest/} This is a large scale chest X-ray dataset. It is labeled with 117 radiological findings - we use the subset with correspondence to the 14 condition labels in the NIH Chest dataset. Images are in 5 different views: posterior-anterior (PA), anterior-posterior (AP), lateral, AP horizontal, and pediatric. 
    \item \textbf{PCAM} \footnote{https://github.com/basveeling/pcam} Patch Camelyon dataset is composed of histopathologic scans of lymph node sections. Images are labeled for presence of cancerous tissue.
    \item \textbf{RIGA} Fundus imaging dataset for glaucoma analysis. Images are marked by physicians for regions of disease. We use this dataset for evaluation 3, use-case 3.
    
\end{itemize}
\section{Details of Experimental Procedure}\label{sec:procedure}
\subsection{Network training}
For classifier models, we use a DenseNet-121 architecture \citep{Huang2017} with Imagenet pretrained weights. The last layer is re-initialized and the full network is finetuned on $D_{tr}$. 
As the NIH and PC-Lateral datasets only contain grayscale images, the pretrained weights of features in the first layer are averaged across channels prior to finetuning. 

For all of the autoencoders, we use a 12-layer CNN architecture with a bottleneck dimension of 512 for all evaluations. Due to computational constraints, all images are downsampled to $64 \times 64$ when fed to an autoencoder. 
These AEs are trained from scratch on their respective $D_{tr}$ with MSE loss and BCE loss. 
We also trained VAEs with the same architectures, except that the bottleneck dimension is doubled to 1024 to allow the code to be split into means and variances.

In addition, we explore the potential benefits of training encoder+decoder using ALI in evaluation 1. We use the same network architecture as proposed in \cite{Dumoulin2016}, with weights pretrained on Imagenet and finetuned on NIH \IN classes. Due to the added complexity of training GANs and the lack of significant improvements in OoDD performance over regular AEs (see \S \ref{sec:results}), we did not train ALI models for the other three evaluations. 

In order to gauge training progress and overfitting, we hold out $5\%$ of $D_{tr}$ as validation set. We select the training checkpoint with the lowest error on $D_{tr}$ for use in OoDD methods.
\subsection{OoDD Method Training}
When training the OoDD methods for use-case 1, three \OUT datasets are randomly selected for $D_{val}$ while the rest is used for $D_{test}$. For use-cases 2 and 3, we enumerate over configurations where each \OUT dataset is used as $D_{val}$ with the rest as $D_{test}$. $D_{val}$ and $D_{test}$ are class-balanced by subsampling equal numbers of \IN and \OUT samples. Additionally, some methods (ODIN and Mahalanobis) require additional hyper-parameter selection. Hence, we further subdivide $D_{val}$ in to a $80\%$ `training' split and a $20\%$ `validation' split; methods are trained/optimized on the `training' split with early-stopping/calibration on the `validation' split. Hyperparameter sweep is carried out where needed. 10 repeated trials, with re-sampled $D_{val}$ and $D_{test}$, are performed for each evaluation. 

\section{Additional Results}
\begin{figure*}[ht]
    \centering
    \vspace{-30pt}
    \includegraphics{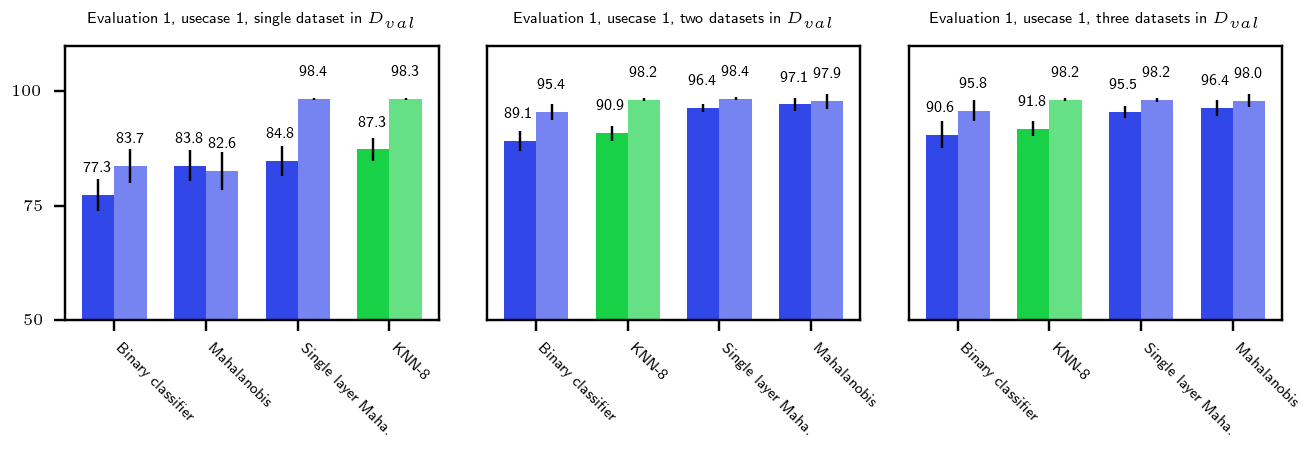}
    \vspace{-25pt}
    \caption{Performance of top-4 methods on frontal X-ray imaging, use-case 1, when trained with fewer datasets in $D_{val}$}
    \label{fig:top4_perf}
\end{figure*}

\end{document}